# COLLABORATIVE CREATION OF DIGITAL CONTENT IN INDIAN LANGUAGES

## QUESTIONS AND ANSWERS

**(This writeup is also available as an HTML file. Contact address at the end.)**

# 1 WHAT IS INFORMATION REVOLUTION?

The world is passing through a major revolution called the information revolution, in which information and knowledge is becoming available to people in unprecedented amounts wherever and whenever they need it. Those societies which fail to take advantage of the new technology will be left behind, just like in the industrial revolution.

The information revolution is based on two major technologies: computers and communication. These technologies have to be delivered in a COST EFFECTIVE manner, and in LANGUAGES accessible to people.

One way to deliver them in cost effective manner is to make suitable technology choices, and to allow people to access through shared resources. This could be done throuch street corner shops (for computer usage, e-mail etc.), schools, community centres and local library centres.

# 2 WHERE ARE INDIAN LANGUAGES IN THE INFORMATION REVOLUTION?

We assume that for a real impact, it is important that the information is available in Indian languages. This requires that certain key technologies and resources should be available in Indian languages. Where are we with respect to these?

## 2.1 DIGITAL CONTENT

A key need is for the digital content (or electronic texts etc.) to be available in Indian languages, tailored to our environment and needs. Presently, very small amount of educational and informative material is available. Most web sites with Indian language content at present have newspapers. The information is of value for a short duration only. Compact disks (CDs) with Indian language content are few, and expensively priced.

## 2.2 MACHINE TRANSLATION

Automatic translation can allow a user to read electronic material from one language in another. Using the current state-of-the-art technology, it is not possible to build fully-automatic, general-purpose, high-quality machine translation system for any pair of languages of the world.

However, it is possible to build language-access systems which allow a reader to read and understand electronic material in another language, provided he is willing to put in some effort. The output of such systems would not qualify to be called as a finished translation, and might also entail some training-to-read on part of the user.

Anusaaraka language access systems have been shown to be feasible from one Indian language to another. (Presently alpha-versions of the anusaaraka systems allow a Hindi reader to read material from five languages namely, Telugu, Kannada, Bengali, Marathi, and Punjabi). It should also be possible to build such systems from English to Indian languages, but a large scale prototype is yet to be built.

## 2.3 SPEECH

Speech processing systems for English are now available commercially, and are likely to become extremely important. They allow the computer to "read out" a given text (text-to-speech), and the computer to understand a spoken utterance (speech-to-text). The latter task is harder, but commercial systems for English have now appeared. Indian language speech technology is nowhere near. Accelerated effort needs to be made if we want to catch up in the next 5 years.

**2.4 OPTICAL CHARACTER RECOGNITION**

Optical character recognition is another technology that allows the computer to "read" a printed page. A printed page on paper can be scanned using a scanner, and its image (like a photograph) can easily be stored in the computer. However, for advantage to be taken regarding search and other language processing tasks (as discussed elsewhere in this writeup - digital content), characters have to be recognized from the image, and the image file is to be converted into a text file. These systems are available and are in routine use for English for past several years. Practical Indian language OCRs seem far away.

**2.5 SEARCH**

Search technology allows a digital library to be searched by the user to locate material needed by him. This technology is in wide-spread use for English, at a massive scale. This technology is relatively easy to build, but requires dictionaries, thesauri, for the concerned language.

**3 HOW CAN I CONTRIBUTE TO THE DEVELOPMENT OF INDIAN LANGUAGE TECHNOLOGY?**

You need not be a computer professional to contribute to Indian language technology development. There are many tasks which can be done by any interested non-experts. For example, content in Indian languages can be entered in the computer by people who can do nothing more than typing on the computer. Other technologies such as machine translation, speech processing, etc. require the preparation of electronic dictionaries, thesauri, other lexical resources, for Indian languages. In fact, the quality of systems produced using these technologies depends crucially on the availability of comprehensive, high quality, electronic, lexical resources. Ordinary people can contribute. An aptitude for language, and sincerity towards work, is all that is needed to contribute towards their creation. Such resources can be created by thousands of such people working in a coordinated manner. It is not necessary that they have to be at one place. Computer networks make this possible. The best model to undertake such a massive effort is to use the voluntary-collaborative model, to create "free" resources, which can be used to build "free" technology for Indian languages available to everybody. ("Free" here means "open", in the sense that people can freely contribute to, modify, and redistribute existing "free" resources, thus leading to their rapid development. See later (Q.14) for further discussion.)

**4 WHAT IS A DIGITAL LIBRARY?**

A digital library is one which has electronic copies of books, rather than paper copies. Such a book is read using computers.  (See digital book.) Clearly the digital library will be a very important part of the information revolution. Such a library would be accessible from any electronic booth, which could be located in schools, community centres, local library centres, etc.

**5 WHAT IS A DIGITAL BOOK?**

A digital book is one in which the material or the content of the book is in electronic form. Such a book is read using computers or similar equipment which displays the content on a screen for the reader. Physically the book could be stored on a compact disk (CD), floppy, magnetic disk, computer memory, etc.

**6 WHAT ARE THE ADVANTAGES OF DIGITAL BOOKS OR E-BOOKS? (Brief answer)**

The major advantages of digital books (namely, electronic books) are: Low cost of reproduction, convenient and more compact storage, easier distribution to users, facility for searching the digital library by the user, organizing the books as hyper-text for easier access to required knowledge, and expert help by the computer.

**7 WHAT ARE THE ADVANTAGES OF DIGITAL BOOKS OR E-BOOKS? (Long answer)**

**7.1 COST**

The major advantage of a book in digital form (namely, books in electronic form) is the lower cost of reproduction, storage and distribution. First, we take the reproduction cost.

For example, a compact disk (CD) can accommodate 500 books (of 500 pages of texts each with 300 words per page) at a reproduction cost of Rs.50/-. This works out to 10 paise per book. (The costs could be higher, if the book contains high-quality pictures as well. On the other hand, if newer technologies such as DVDs are used, the cost becomes lower.)

Internet also makes the distribution of digital books easier. Although, the bandwidth costs turn out to be higher, it has the advantage of the updated material available all the time. Thus, a judicious combination of CDs and web based material can be used with great cost effectiveness.

**7.2 CONVENIENT STORAGE**

Storage of digital books is easier than paper as they take much less space. A CD containing 500 books (of text only) fits in a shirt pocket.

**7.3 DISTRIBUTION EASIER**

Distribution costs of digital books are also low. CDs can be distributed to libraries or a centre in each city, from where they can be accessed by the readers. The final access could be through a CD in a local machine or by using ordinary telephone lines to a nearby or remote centre having the digital book.

Essentially, the reproduction, storage and distribution cost of a book is approaching zero.

**7.4 SEARCH**

A digital book can be used in new ways not possible with the paper copies of books. The computer can perform search in the digital library to help a user locate the information he wants rapidly. Powerful search engines are available that try to locate the information wanted by a user rapidly.

**7.5 HYPER-TEXT**

Digital books are increasingly being based on hyper-text which present the topics in a book, in different ways to the reader according to his requirements. Thus, the digital books need not be in linear format like the paper books.

### 7.6 EXPERT SYSTEM

Computer software (called expert systems) will be able to use the information in digital libraries to perform useful tasks for us in the future.

### 7.7 DETAILS DISCUSSED

There are many issues in the above. However, the following are taken up in detail in questions below. (Other issues are not taken up here.)

(i). Hyper-text

(ii). Cost of equipment or computers to read the book.

(iii). Creation of content (author's effort and intellectual property rights)

### 8 WHAT IS HYPER-TEXT ?

A digital book can be so organized that it can be read in different ways based on topics, levels of detail needed, etc. For this the author has to connect relevant bite-sized chunks such as paragraphs, sections, etc. together, anticipating different uses. Links which connect different chnnks together, are called hyper-links. Thus, the book is not just a linear sequence of pages, but one in which there are many connections. This is called the Hyper text.

A computer allows the readers to follow different links of their choosing in the hyper text easily. For example, if you are reading this document in hyper-text form, you can jump across questions, get a list of questions, etc. More hyper links can be added to it depending on the needs of different readers.

### 9 ARE COMPUTERS AFFORDABLE? (Brief answer)

The cost of computers or equipment to access digital content is high, say around Rs.40,000. (including PC hardware (Rs.30,000) and commonly used PC software (Rs.10,000)).

Innovative solutions are available to lower the cost. These include the use of thin clients (such as Indian script GIST terminals), and the use of free-software such as Linux and GNU (as opposed to Windows-98 and Office 97). Such a text terminal needs a host PC to operate from, which could be located nearby or be remotely accessible through the network or telephone.

The cost of the above (thin client) computer terminal is around Rs.11000. The cost of a computer seat, therefore, works out to be around Rs.13000., if the shared cost of the host PC is included. (This cost includes software not just for internet, email, word processing and spread sheet, but also programming languages, software tools, DBMS, etc.)

The cost can go down further to Rs.8000. or less, if thin clients are manufactured on a large scale. What is needed is awareness regarding such solutions and willingness to learn to use them.

Assuming the cost of a seat as Rs.13,000., the recurring expenses (covering maintenance, interest on loan, depreciation) turn out to be Rs.6,500. per year. This works out to Rs.20. per day. If the machine is used for 10 hours, this turns out to be, around Rs.2. per hour. To this can be added the cost of communication in roughly the same range.

Some believe that it is possible to reduce the cost of a terminal to Rs.5000. by clever engineering. Or if the user is willing to use his TV as a display device, and a low cost (less comfortable) keyboard, the cost can go down to Rs.2000. This does not seem impossible. As we have seen in other areas (see Jhunjhunwala's article), the imperative to reduce the cost goes down in the West, once cost reaches a certain level. The challenge of further reducing the cost needs to be undertaken by engineers in developing countries to suit our requirements.

**10 ARE COMPUTERS AFFORDABLE? (Long Answer)**

The cost of computers or equipment to access digital books is high. A typical low cost computer which can operate independently costs around Rs.30,000/- in hardware and an additional Rs.10,000/- for some basic software (for email, word processing, spreadsheet, with say, Windows-95.) It is also difficult to share, as one user can tamper with the work done by another.

The key to reduce the cost is to :

1. Use innovative hardware-software combination.

2. Share the machine with several users (over time).

10.1 Innovative Solutions

Innovative solutions are available to lower the cost. These reduce the cost dramatically, if a number of "computer seats" are to be provided in a room or a building. The hardware cost of an independent computer (PC) would be around Rs.30,000/-. To this computer, one could connect, say, 8 thin clients such as Indian script GIST terminals at a price of Rs.11,000/- each. The total cost (Rs.1,18,000) divided by 9 "computer seats" (8 GIST terminals and one PC) turns out to be Rs.13,000/- per computer seat. (The cost can be reduced further to say Rs.8,000/- if the thin clients are manufactured on a large scale).

Free-software such as LINUX and GNU (as opposed to Windows-98) can be used to support the above configuration. (Windows-98 would cost money, besides it does not have the capability to support such a configuration.)

It is important to realize the quality and power of free software. It not only includes software for internet, email, word processing, spread sheet but also includes a host of programming language compilers, software tools, database management, etc. Price of buying such software for Windows-98 would run into a lakh of Rupees. Not only this, the free software is usually faster and uses less memory space than the priced versions of such software for Windows-98. Most important of all, source code is "open" and can be tinkered with, to suit our requirements.

With GIST terminals, all display and keyboarding capability in all Indian languages, comes as part of the hardware itself. As we have seen earlier, these reduce the cost per seat substantially. The limitation is that these are text-only seats. However, there would be one graphics seat in the above configuration namely the console on the PC. Some amount of learning will also be needed because people are not familiar with free software at present.

**10.2 Sharing the Use of Computers**

The above solution (rather than Windows-98) is also designed for shared use. It allows the same CD on a CD drive to be shared among several readers at the same time. With a single dialup phone connection, internet and email becomes available on every seat. This lowers the phone connect charges substantially. It also allows users to save their work on disk which is protected from other users modifying it (or even reading it, if so decided). There is no such protection on Windows-98. Thus, a computer seat can be shared among different users over time without the fear of their affecting each others work. Because of such protection, these machines are also relatively immune from computer viruses.

If we assume the cost of a computer seat as Rs.13,000/-, the annual recurring expenses (maintenance 10%, interest on loan 15%, depreciation 25%) turn out to be Rs.6,500/- per year. This works out to Rs.550/- per month or Rs.20/- per day. If the machines can be put to use for 10 hours a day (27 days a month), the cost is Rs.2/- per hour.

**10.3 Future Cost Projections**

Some believe that it is possible to reduce the retail cost of a terminal to Rs.5000. by clever engineering. (Or if the user is willing to use his TV as a display device, and a low cost keyboard, the cost can go down to Rs.2000.)

**11 HOW CAN A DIGITAL BOOK BE CREATED ?**

Digital books can be created by keying in a book into the computer provided the copyright permits it. Hyper links can be created within it to allow access in easier chunks or to permit a specific topic or issue to be pursued by the reader. Of course, the new books can be created directly in this form by their authors.

12 HOW DOES THE NEW TECHNOLOGY AFFECT THE AUTHORING OF A DIGITAL BOOK?
   HOW DOES THIS PROCESS DIFFER FROM THAT OF A PAPER BOOK?

When a manuscript is in electronic form, two things are easy: First, it can be modified very easily. Second, it can be transmitted to a collaborating author at very low cost.

This permits many people to work together in creating a digital book. One person writes a part of the book and sends it to his colleague who gives feedback, fills in the missing parts etc. and sends it back easily, for a similar reaction from the first person. Or alternatively, they might seek another expert to fill in a missing part. All this leads to a highly collaborative style, resulting in, better quality and more authoritative books.

As the cost of distribution is low, many people can become authors. Thus, there are likely to be many more books created. all might be of good quality though). However, out of these books will emerge many more high quality books than at present.

A new process seems to be in motion with the pervasive network and e-mail in which thousands of people might contribute to the creation of a single book. An electronic book available over the internet might be improved in myriads of ways by thousands of readers, the updated edited version becomes available to everybody. It would require changes in intellectual property rights so that they do not stop somebody from making better quality books available to everybody. In fact, this brings back the importance of cooperative models rather than the private models.

(Certain kinds of books particularly literature would continue to be produced by individual authors.)

## 13 WHAT IS THE STATUS OF INDIAN LANGUAGE CONTENT IN ELECTRONIC FORM?

There is an acute paucity of material in Indian languages in the electronic form. CDs available are few, and the internet also has very little material. If our people have to take benefits of the information society and India as a nation has to prosper, it is extremely important that knowledge becomes available in Indian languages in the electronic form.

## 14 HOW CAN WE CREATE ELECTRONIC CONTENT IN INDIAN LANGUAGES ?

The first thing that is required is that people connected to education and research should start creating content in their respective subject areas in Indian languages in electronic form.

If we choose the free-software model for electronic content, the whole process can be speeded up. In this model, we contribute our labour creating digital content. At the simplest level people can enter in the computer, existing published material whose copyright has expired or after obtaining permission for entering it. People can also create a new book or free-text on the internet, and give permission to others to not only read it freely, but also modify it and place the new versions as free-text (with the same conditions). This permits and encourages thousands of readers to contribute their labour. One reader might help in proof-reading, another might build hyper-links, yet another might refine the material further. Editorial work would also be needed to control and coordinate the refinements being done by myriads of people, but that could again be distributed in a similar way albeit a little more carefully. A task that is highly suitable for creation through such work are mono-lingual and bilingual dictionaries, thesauri, etc.

The presence of software tools such as anusaaraka (a kind of machine translation software) that allow access across Indian languages can also help in making content in one Indian language available in another Indian language. The output from such tools can be further post-edited by people to produce translations and kept over the net.

## 15 CAN THE COLLABORATIVE METHOD FOR CREATION OF DIGITAL CONTENT WORK?

It is said that if everybody were to write parts of a book, there would be no continuity, besides repetitions and other problems in the book. The quality of books would be low. The answer turns out to be different from what one might expect to happen.

Let us take the case of software writing. In the case of software, a single mistake in any one of the modules can render the whole software totally unusable. However, in spite of such sensitivity to mistakes, free software continues to be written by myriads of programmers working together, and the greatest surprise is that it has turned out to be of higher quality than software written under companies under tight control. How has this come about? The number of people willing to read and test computer programs is very large, as a result errors in such open software get caught and corrected much faster than those written by companies. Of course, this requiresa degree of coordination and control, which is brought about by the community itself without a tight central control. A single person coordinates, but he gets free-helping hands, and most importantly, when the coordinator wants to move on to other things, his coordinated software is taken over by another and continues to live.

If software can be written by such a multitude of authors, certainly books and texts can be written this wway, and they are likely to b superior to other modes of their creation. They will also tend to get refined and updated even after the initial author(s) has moved on to other things.

To create digital content in our languages, the fastest way to move forward and involve thousands of people in its creation is to adopt the cooperative-collaborative model. No company can match the speed (and ultimately quality) in this model.

## 16 WHAT CAN I DO? WHAT ARE SOME OF THE CONCRETE TASKS THAT CAN BE DONE COLLABORATIVELY?

If you are willing to donate your labour for the development of digital content in Indian languages, a number of tasks can be undertaken. Some examples are given below.

### 16.1 Selection:

You can help in selection of existing material which is of high quality and should be available in digital form. (List of such selected materials can be put in a central repository on the Indian internet.)

### 16.2 Seeking permission:

In case the copyright of the selected material is held by somebody, he or she can be approached for permission to enter the material into computers. You can help in obtaining such permission. (A large number of publishers who hold copyright to scholarly materials, and whose charter is to disseminate such material rather than to make money, should have no serious problem in giving such permissions. Granth Akademies in different states are public funded bodies with this charter. There are many other such bodies.)

### 16.3 Entering texts:

You can help in entering the material into the computer when either the copyright has expired (in case of old and classic material), or permission has been given by the copyright holder to do so.

### 16.4 Proof-reading:

You can help in proof-reading material that has been typed into the computer.

### 16.5 Hyper-text creation:

If you wish to create material in hyper-text form, you can take existing electronic materials and break them into "bite-size" chunks, and link them by creating hyper-links between them.

### 16.6 Authoring digital books:

If you are an author or wish to be one, you can try to create your material in digital form, and make it available to others for general use, giving them permission to make modifications for further refinement. You can yourself create it in hyper-text form, of seek assistance of volunteers who can take your material and with your help create hyper-links.

**16.7 Editing:**

You can help in editing material that is being written by multiple authors. Editing is also needed when existing material is "refined" by people, before such refined material is accepted and put in the electronic repository.

**16.8 Copy-editing:**

If you are good with language, you can help in copy-editing material written by subject experts. You can make their material easier to read, correct it for grammar, style, etc.

**16.9 Dictionary refinement:**

As a first task, the following five dictionaries are being placed on the internet for free download and refinement: Telugu to Hindi, Kannada to Hindi, Marathi to Hindi, Punjabi to Hindi, Bengali to Hindi. You can go through them, use them for your own purpose, as well as send us feedback about it for its correction, refinement etc.

**16.10 Dictionary refinement -  English to Hindi (current activity):**

As another concrete activity, dictionary from English to Hindi is being created using the collaborative model. If you wish to participate in the task, contact the Akshar Bharati group at  sangal@iiit.net  (Contact address given at the end.)

**17 WHAT ARE THE IMMEDIATE TASKS BEING UNDERTAKEN?**

Creation and refinement of on-line and open dictionaries is being undertaken. (See the previous question.) You can participate in it from your place of work or home if you have access to computers  and email. Contact us at the address given at the end.

**18  WHAT DO I GAIN BY PARTICIPATING IN COLLABORATIVE ACTIVITY?**

By participating in this collaborative activity and creating "free" content, you gain in several ways. First and foremost, you  become part of a process which helps in preparing the nation and  our languages for the next millenium. Second, if you make excellent contibutions you become known and get recognized by the community. Third, you acquire skills by participating in this activity, which will help you become entrepreneur, free lance person, or get jobs.

You might have heard of a story on heaven and hell. A person was taken to a visit to hell. It was lunch time, and much to his surprise, excellent food was being served. But soon there was much noise and commotion, and most of the food was spilled, with most people going hungry. The reason was that on everybody's hands were tied long spoons of several feet. As a result, the people could not eat without tossing the food in their mouths.

He then visited heaven, where the meal had just ended. The food was the same, and the people seem to have just finished a quiet and fulfilling meal. To his utter surprise, here too long spoons were tied to the hands of people. On enquiring, he learnt that the only difference was that here people had fed each other rather than trying to eat by themselves.

Collaborative creation of digital content is a similar activity. Either we can all starve and be frustrated, or we all work together in creating new resources to be used by everybody.

**19 CONTACT ADDRESS**

Akshar Bharati group (Attn: Prof Rajeev Sangal),
Language Technologies Research Centre
International Institute of Information Technology
Gachibowli, Hyderabad 500 019

Tel: +91-40-2300 1412 or 23001967 Ext 144
Fax: +91-40-2300 1413
email: sangal@iiit.net